%% file: root.tex
\PassOptionsToPackage{table}{xcolor}
\documentclass[letterpaper, 10 pt, conference]{ieeeconf}  

\IEEEoverridecommandlockouts                              

\usepackage{graphicx}
\usepackage{subcaption}
\usepackage{amsmath}
\usepackage{mathtools}
\usepackage{amsfonts}
\usepackage{enumerate} 
\usepackage[font=footnotesize]{caption}

\usepackage{pgfplots}
\pgfplotsset{compat=1.18} 

\usepackage{algorithm}
\usepackage{algpseudocode}
\usepackage{comment}
\usepackage{capt-of}
\usepackage{caption}
\usepackage{bm}
\usepackage{eso-pic}

\usepackage{multirow}
\usepackage{xcolor}
\usepackage{makecell} 

\newcommand{\changliu}[1]{\normalsize{\color{blue}(\textbf{CL:}\ #1)}}

\title{\LARGE \bf
Whole-Body Safe Control of Robotic Systems \\ with Koopman Neural Dynamics
}

\author{ 
        Sebin Jung\textsuperscript{1},
        Abulikemu Abuduweili\textsuperscript{1},
        Jiaxing Li\textsuperscript{1},
        Changliu Liu\textsuperscript{1}
}

\begin{document}

\twocolumn[{%
\renewcommand\twocolumn[1][]{#1}%
\maketitle

\begin{center}
\vspace{-3.5mm}
\includegraphics[width=0.5775\columnwidth]{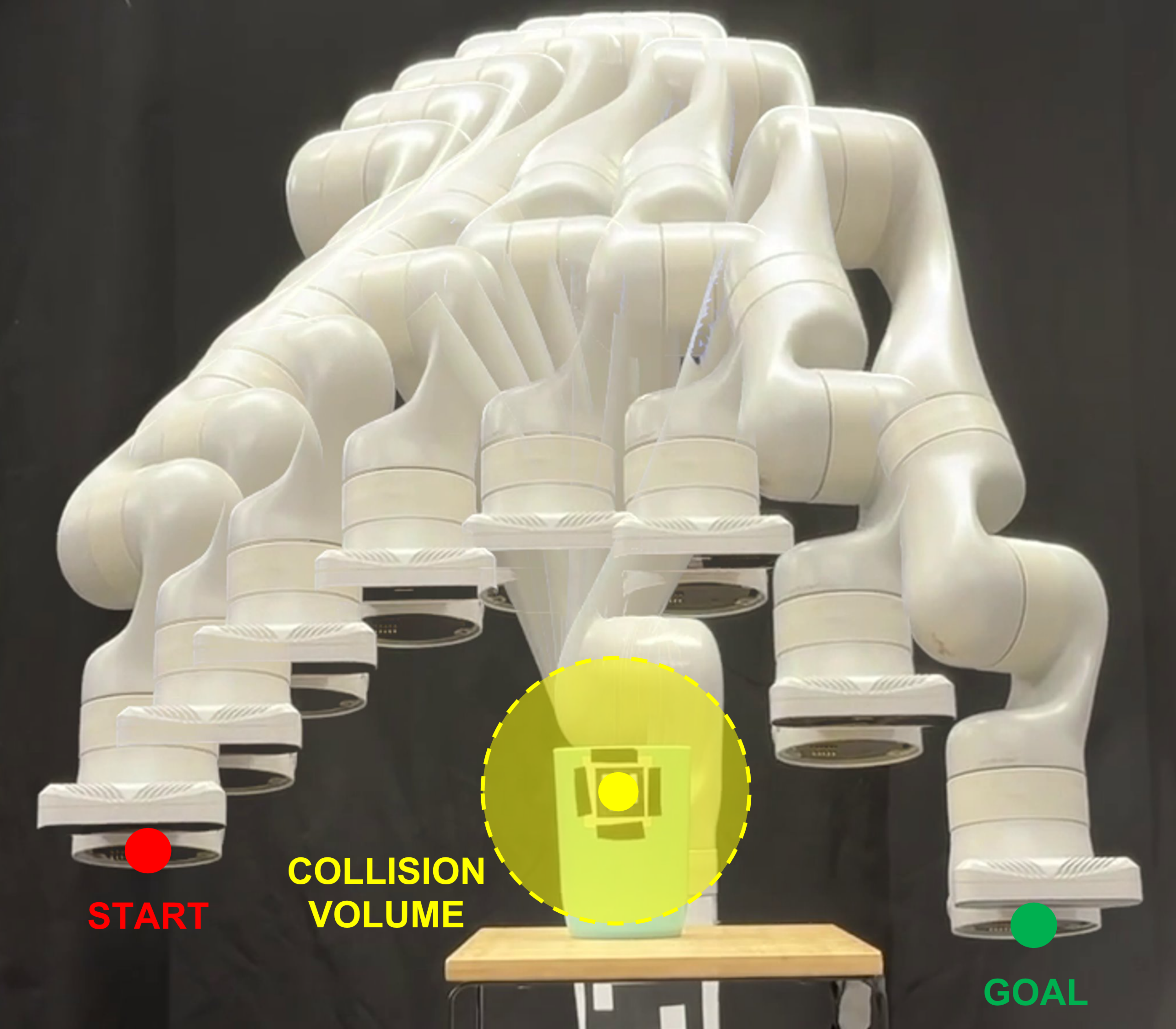}
\includegraphics[width=0.515\columnwidth]{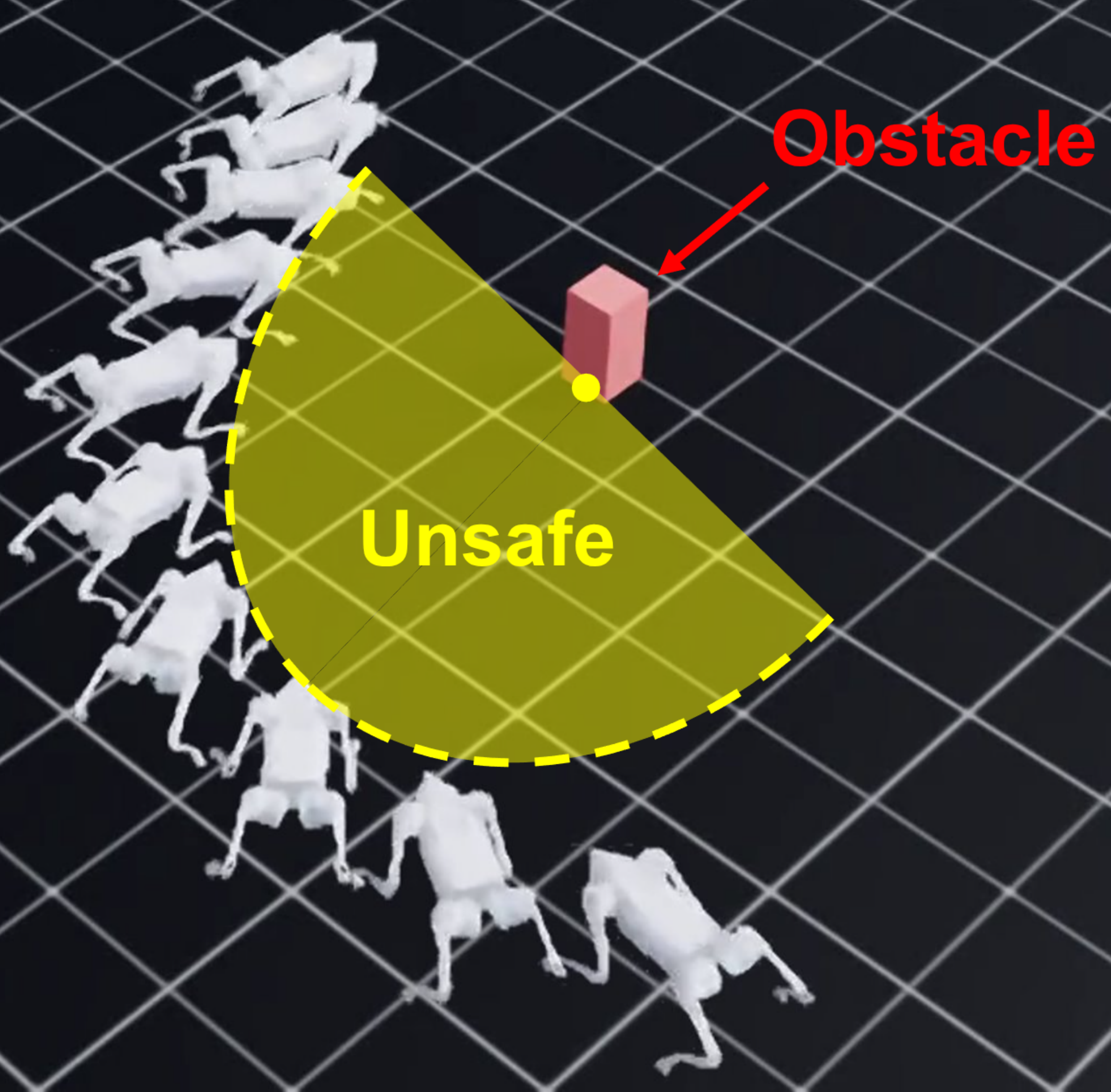}
\includegraphics[width=0.5\columnwidth]{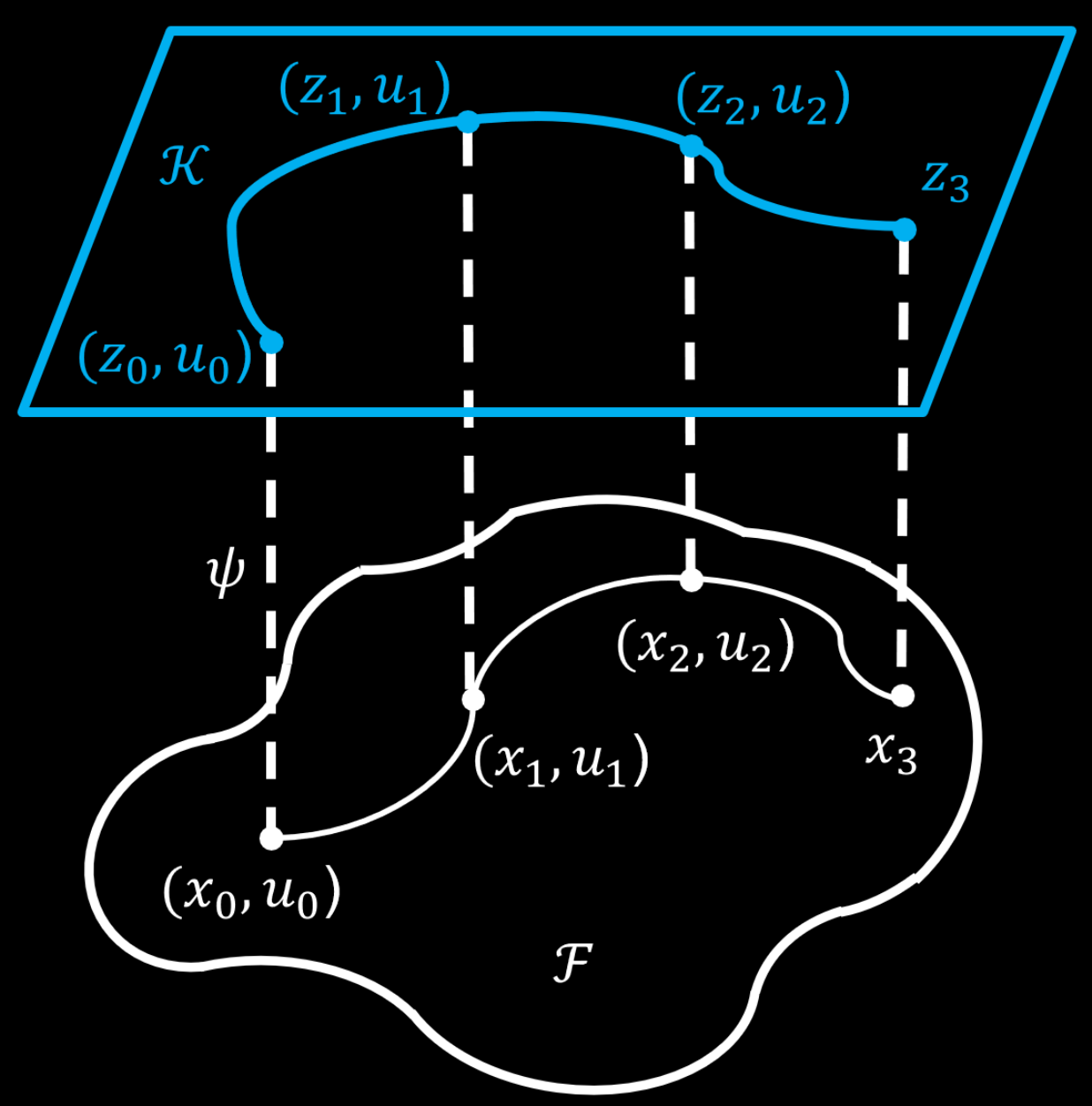}\\[0.8mm]

{\captionsetup{justification=raggedright,singlelinecheck=false,font=footnotesize,skip=2pt}%
\captionof{figure}{\setlength{\baselineskip}{0.92\baselineskip}%
Left: Kinova Gen3 safely moving from START (red) to GOAL (green) while avoiding the collision volume (yellow); collision avoidance is enforced using the learned Koopman dynamics within our unified Koopman--SSA MPC.
Middle: Floating-base safe control in Isaac simulation (Unitree Go2). The learned Koopman-MPC tracks the reference while actively avoiding the obstacle-induced unsafe region; collision avoidance is enforced using the learned Koopman dynamics within the MPC rollout.
Right: We use the learned Koopman lifted space and its linear dynamics (blue) to predict motion in MPC and evaluate the safety constraints, which enables the collision avoidance shown in the left and middle panels.}
\label{fig:title-fig}}
\end{center}

\vspace{-0.1mm}
}]  

\begingroup
\renewcommand\thefootnote{}
\footnotetext{\textsuperscript{1}Robotics Institute, Carnegie Mellon University, Pittsburgh, PA. Contact: \texttt{\{sebinj, abulikea, jiaxingl, cliu6\}@andrew.cmu.edu}}
\endgroup

\addtocounter{figure}{1}  

\begin{abstract}

Controlling robots with strongly nonlinear, high-dimensional dynamics remains challenging, as direct nonlinear optimization with safety constraints is often intractable in real time. The Koopman operator offers a way to represent nonlinear systems linearly in a lifted space, enabling the use of efficient linear control. We propose a data-driven framework that learns a Koopman embedding and operator from data, and integrates the resulting linear model with the Safe Set Algorithm (SSA). This allows the tracking and safety constraints to be solved in a single quadratic program (QP), ensuring feasibility and optimality without a separate safety filter. We validate the method on a Kinova Gen3 manipulator and a Go2 quadruped, showing accurate tracking and obstacle avoidance.

\end{abstract}

\input{sections/1_intro}

\input{sections/2_preliminaries}
\input{sections/3_problem_formulation}

\input{sections/4_methodology}

\input{sections/5_experiments}

\input{sections/6_future_work}

\input{sections/7_conclusion}

\bibliographystyle{IEEEtran}
\bibliography{bibtex/ref}

\clearpage
\newpage

\appendix
\input{sections/8_appendix}

\end{document}

%% file: sections/1_intro.tex

\section{INTRODUCTION}

Ensuring safe control of robotic systems is fundamentally difficult due to the interplay of complex dynamics, high dimensionality, and safety-critical constraints. A first challenge arises from the strong nonlinearity of the dynamics: joint couplings, contact interactions, and nonlinear actuation make the direct embedding of such models into optimization-based controllers computationally prohibitive. Even when accurate models are available, nonlinear formulations with safety constraints often lead to nonconvex programs that cannot be solved in real time \cite{gros2020linear}. A second challenge is on feasibility, which emerges at the boundary of the safe set, where a controller may fail to generate feasible inputs to steer the system back into safety. Finally, when using learned dynamics, approximation errors can propagate into safety constraints, rendering the assumed safe control unsafe. 

A range of data-driven methods learn predictive models of nonlinear dynamics, but they rarely translate into real-time safe control.
Dynamic Mode Decomposition and its extensions approximate nonlinear systems with linear operators in lifted spaces~\cite{schmid2010dynamic,williams2015data,korda2018linear}, while sparse-regression methods such as SINDy identify compact governing equations~\cite{brunton2016discovering}.
Deep recurrent and latent-state models extend predictive power~\cite{watter2015embed}, and Koopman operator theory provides a principled framework for globally linearizing nonlinear dynamics~\cite{feihan2025Koopman,lusch2018deep,chen2025korol}.
Yet these approaches focus on prediction, and do not ensure that modeling errors or online computation limits can meet hard safety constraints during control. 

On the control side, no existing family of methods simultaneously handles high dimensionality, model uncertainty, and strict safety guarantees.
Classical nonlinear controllers such as feedback linearization require exact models~\cite{khalil2002nonlinear}, and trajectory-optimization techniques like Iterative Linear Quadratic Regulator (iLQR) or Nonlinear Model Predictive Control (NMPC) scale poorly as degrees of freedom grow~\cite{tassa2012synthesis}. 
Reinforcement learning has shown striking successes in robotics~\cite{schulman2017proximal}, but it demands large data and provides only probabilistic safety at best~\cite{cheng2019endtoendsafereinforcementlearning}. 
Control Barrier Functions (CBFs) and Safe-Set Algorithms provide convex safety-filtering mechanisms~\cite{ames2016control,wei2022safe} yet remain difficult to formally synthesize with learned or highly nonlinear models in real time. Even when synthesis is learned from data, optimization-based coupling with hard safety constraints is challenging and often exhibits feasibility issues in practice. Moreover, under strongly nonlinear dynamics, safety is frequently enforced via a separate filtering layer on top of a nominal controller, which can induce performance--safety trade-offs such as overly conservative behavior or deadlock. Recent efforts to merge Koopman embeddings with CBFs~\cite{zinage2022neuralkoopmancontrolbarrier} still separate nominal control from safety filtering, creating feasibility breakdowns near the boundary.

\begin{figure*}[h]
  \centering
  \includegraphics[width=\textwidth]{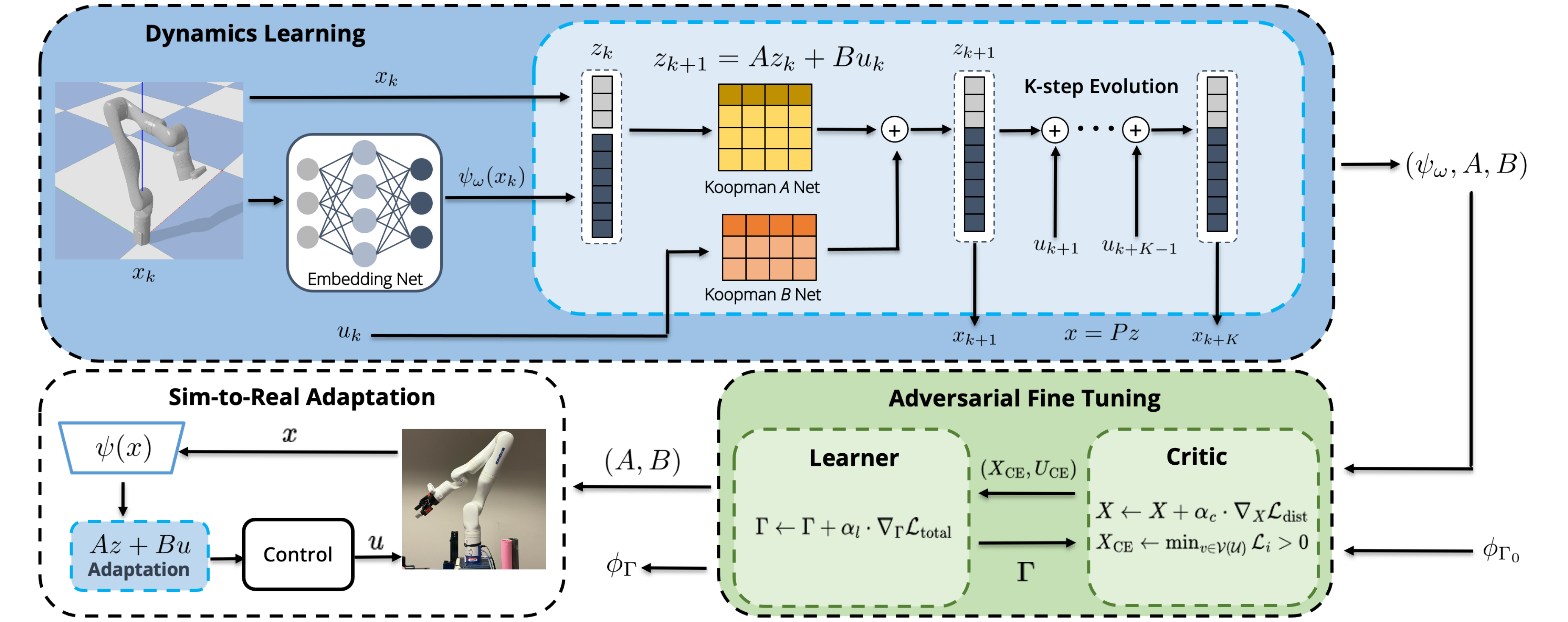}
  \caption{Overview of training framework for Koopman Safe Control. Neural embedding function and Koopman Operators are first trained, safety index is adapted to the learned dynamics via adversarial fine tuning, and the model is migrated to real environment.}
  \label{fig:overview}
\end{figure*}

We propose a data-driven safe control framework that combines Koopman operator theory with whole-body robot dynamics. This framework (Fig. \ref{fig:overview}) tackles nonlinearity, safety-boundary infeasibility, and learned-model approximation errors, enabling scalable, real-time safe control of nonlinear robotic systems. The main contributions are:

\textbf{Synthesis of Safe Control via Koopman Linearization.} We formulate whole-body safe control of robotic manipulators by globally linearizing their nonlinear dynamics through the Koopman operator. Unlike filtering- or projection-based two-stage architectures \cite{wabersich2021predictive}, which apply safety constraints as a post-processing step on nominal controls, our approach synthesizes safety directly into the nominal controller. This unified formulation ensures consistent reasoning about performance and safety while avoiding the conservatism and inefficiencies of safety filtering.

\textbf{Safety Index Synthesis for Learned Dynamics.} To mitigate infeasible control issues that arise when coupling learned Koopman dynamics with hard safety constraints, we introduce an adversarial fine tuning scheme for the safety index. This procedure adapts the safety specification to the learned dynamics, improving forward invariance inside the safe set by significantly reducing the risk of unsolvable constraints.

\textbf{Adaptation to Real-World Hardware.} We demonstrate the effectiveness of the proposed framework through sim-to-real deployment on a Kinova Gen3 manipulator with minimal retraining. As illustrated in Fig. \ref{fig:title-fig}, the Kinova safely tracks the given trajectory while avoiding the designated collision volume (yellow). A video demonstration of this experiment is provided in the supplementary material. 

%% file: sections/2_preliminaries.tex
\section{PRELIMINARIES}

\subsection{Koopman Operator Theory}

Given the nonlinear system with discrete dynamics: 
\begin{equation}
x_{k+1} = f(x_k)
\label{eq:nonlin_sys}
\end{equation} 
Koopman Operator $\mathcal{K}$ is an infinite dimensional linear operator that maps the nonlinear dynamics to linear dynamics \cite{brunton2021modern} and evolves the embedding functions $\psi$ of the state: 
\begin{equation}
\mathcal{K}\psi(x) = \psi \circ f(x)
\label{eq:evolve}
\end{equation} 
Considering a nonlinear system with the control input $u_k \in \mathcal{U} \subseteq\mathbb{R}^m$ and system state $x_k \in \mathcal{X}\subset\mathbb{R}^n$, given by $x_{k+1} = f(x_k,u_k)$, the evolution flow follows as: 
\begin{equation}
\mathcal{K}\psi(x_{k},u_{k}) = \psi(f(x_{k},u_{k})) = \psi(x_{k+1})
\label{eq:evol_flow}
\end{equation} 
where $\psi: \mathbb{R}^{n+m} \rightarrow \mathbb{R}^{\infty}$. In practice, we constrain the latent space to a finite-dimensional vector space and approximate the Koopman operator, making $\psi: \mathbb{R}^{n+m} \rightarrow \mathbb{R}^{d}$ where $d\in\lbrack0,\infty)\subset\mathbb{R}^{+}$. 

The embedding function $\psi(x, u)$ can be separated as $\psi(x, u) = \lbrack \psi_x(x, u); \psi_u(x, u) \rbrack$, and we focus on the case where $\psi_x(x,u) = \psi_x(x)$ and $\psi_u(x,u)=\hat{u}=u$. Here $\psi_x: \mathbb{R}^n\rightarrow\mathbb{R}^{n^\prime}$ denotes state embedding and $\hat{u}$ is control in latent space. Under this formulation, the system dynamics can be expressed as: 
\begin{equation}
    \psi_x(x_{k+1}) = A\psi_x(x_k) + Bu_k
    \label{eq:latent_dyn}
\end{equation}
where $A,B$ comes from:
\begin{equation}
    \mathcal{K} = \begin{bmatrix}
        A\in\mathbb{R}^{n^{\prime}\times n^{\prime}} & B\in\mathbb{R}^{n^{\prime}\times m} \\ 
        C\in\mathbb{R}^{m\times n^{\prime}} & D\in\mathbb{R}^{m\times m}
        \end{bmatrix}
        \label{eq:lifted_k}
\end{equation}

We parameterize state embedding $\psi_x$ as a neural network $\psi_{\omega}$ and denote the lifted state $z_k = \psi_x(x_k) = \lbrack x_k ;\, \psi_{\omega}(x_k) \rbrack^\top$. This preserves the original state data along with the neural network embedded state in order to prevent the information loss that may occur through nonlinear mapping~\cite{shi2022deep}. Specifically, we can retrieve the original state via:
\begin{equation}
    x_k = Pz_k 
    \label{eq:retrieve}
\end{equation}
where $P = [I_n;\,0] \in \mathbb{R}^{n \times (n+n^\prime)}$. Then, the lifted state evolution takes the form:
\begin{equation}
    z_{k+1} = Az_k + Bu_k
    \label{eq:lifted_evol}
\end{equation}
This formulation allows us to design linear controllers for state-constrained control problems (e.g., collision avoidance) by leveraging the linearity of lifted dynamics. In addition, we use end-to-end training framework for the acquisition of embedding function as well as the Koopman operator $(\psi_\omega, A, B)$. Readers can refer to \cite{brunton2021modern} for further understanding of Koopman Operator Theory.

\subsection{Safe Control Synthesis through Adversarial Fine Tuning}

The goal of safe control is to design a controller that tracks a reference while ensuring forward invariance inside the allowable set $\mathcal{A}$ \cite{liu2014control}. For a function $s: \mathbb{R}^n\rightarrow \mathbb{R}$ let $\mathcal{S} \coloneq \{s\}_{\le0}$ be its zero-sublevel set, $\partial\mathcal{S} \coloneq \{s\}_{=0}$ the boundary of this set. Assuming we are given: (1) a control-affine system $\dot x = f(x) + g(x)u$ where $f:\mathbb{R}^n\rightarrow\mathbb{R}^n$ and $g:\mathbb{R}^n\rightarrow\mathbb{R}^{n\times m}$ are locally Lipschitz continuous on $\mathbb{R}^n$, (2) the control set $\mathcal{U}$ is a bounded convex polytope, (3) safety specification $\phi_0: \mathbb{R}^n\rightarrow\mathbb{R}$ implicitly defines the allowable set $\mathcal{A}\coloneq\{\phi_0\}_{\le0}$.

If we directly impose the constraint $\dot\phi_0(x) \le 0$ at the boundary of allowable set $\partial \mathcal{A}$ for a limit–agnostic index $\phi_0$ and bounded $\mathcal{U}$, the feasible set can collapse, causing controller saturation or an unsolvable constraint. We therefore \emph{reshape} the safety specification by introducing a parameterized index:
\begin{equation}
    \phi_{\rho} = h(\rho, \phi_0)
    \label{eq:learnable_phi}
\end{equation}
where $h:\mathbb{R}\rightarrow\mathbb{R}$ with learnable parameters represented as $\rho$. The zero-sublevel set of $\phi_{\rho}$ defines a more conservative safe set $\mathcal{S}:=\{x|\phi_\rho(x)\leq 0\}$. At its boundary, the safe control constraint is defined as:
\begin{IEEEeqnarray}{rCl}
    \dot\phi_{\rho}(x,u) = \nabla\phi_{\rho}(x)^{\top}f(x) + \nabla\phi_{\rho}(x)^{\top}g(x)u \le 0 \nonumber\\
    \forall x \in \partial{\mathcal{S}}
    \label{eq:dot_phi}
\end{IEEEeqnarray}
in order to guarantee forward invariance inside $\mathcal{S}$. Readers can refer to \cite{pmlr-v205-liu23e} for further details.

To ensure feasibility under input limits and model approximation, we adapt $\rho$ via the min–max program:
\begin{equation}
\min_{\rho}\;\max_{x \in \partial\mathcal{S}}\;\inf_{u \in \mathcal{U}} \dot{\phi}_{\rho}(x,u)
\label{eq:adv_obj}
\end{equation}
which can be evaluated over the vertex set of control $v \in\mathcal{V}(\mathcal{U})$ when $\mathcal{U}$ is a polytope \cite{pmlr-v205-liu23e}. Thus, the objective can be computed as a discrete minimization:
\begin{equation}
\mathcal{L}(\rho,x) \coloneq \min_{v \in \mathcal{V}(\mathcal{U})} \dot\phi_{\rho}(x,v)
\end{equation}
This procedure \emph{learns} a safety index that avoids infeasible constraints while preserving forward invariance.

%% file: sections/3_problem_formulation.tex
\section{PROBLEM FORMULATION}

In this paper, we consider whole-body safe control of an $n$-DoF articulated rigid-body system (including fixed-base manipulators and floating-base legged or wheeled robots) operating amid static and dynamic obstacles. The target system is the dynamics in generalized coordinates:
\begin{equation}
M(q)\ddot{q} + C(q,\dot{q})\dot{q} + G(q) = \tau + J^\top (q)f
\label{eq:manip_dyn}
\end{equation}
where $q \in \mathbb{R}^n$ denotes a vector of  generalized coordinates (e.g., joint positions for fixed-based robots; joint and base positions for floating-based robots), and $\dot{q}$ the corresponding generalized velocities. Here $M(q)$ is the inertia matrix, $C(q,\dot{q})$ the Coriolis and centrifugal terms, $G(q)$ the gravity vector,  $\tau$ the applied joint torques, $J (q)$ the contact Jacobian, and $f$ the contact forces. Equation~\eqref{eq:manip_dyn} represents the low-level dynamics that must ultimately be respected, but which are highly nonlinear and high-dimensional to directly embed into real-time safety-critical controllers. 

Our control objective is trajectory tracking in the presence of obstacles, under uncertainty in both robot motion and obstacle motion. We consider all obstacles and robot link collision volumes to be spheres. This choice allows (i) distance functions and gradients admit closed-form expressions, enabling efficient evaluation of safety constraints, and (ii) spherical volumes approximate local link geometry while simplifying multi-link safety verification into a minimum-distance calculation between spheres.

We define a signed distance from the ego towards the obstacle as safety specification. Let the state be $x:=q$. For a target link's center of mass $p_{\text{ego}}(x)\in \mathbb{R}^3$ and obstacle center $p_{\text{obs}}\in \mathbb{R}^3$:
\begin{equation}
\phi_0(x) = d_{\min} - \| p_{\text{ego}}(x) - p_{\text{obs}} \|_2
\label{eq:safety_spec}
\end{equation}
where $d_{\min}\in \mathbb{R}^{+}$ is the minimum distance to be maintained between ego and obstacle. The nonlinear state-based safety specification $\phi_0(x)\le 0$ can be posed as state-dependent constraints on the control space:
\begin{equation}
\dot{\phi}_0(x_k,u_k) \le b_\phi(x_k)
\label{eq:safety_constraint}
\end{equation}
where the safety bound $b_\phi(x)$ is defined as
\begin{equation}
b_\phi(x) =
\begin{cases}
0 & \text{if } x \in \partial\mathcal{A}\\
-\lambda & \text{if } x \notin \mathcal{A}\\
\infty & \text{otherwise}
\end{cases}
\label{eq:bphi}
\end{equation}

This safe control constraint can be directly embedded into NMPC.
However, solving with full nonlinear dynamics is computationally prohibitive and often infeasible in real time, while local linearizations quickly lose validity away from the expansion point. Synthesizing safety-filtering methods is nontrivial especially at the safe set boundary, and black-box learned models risk false safety guarantees due to approximation error. These drawbacks highlight the need for an approach that embeds safety directly while remaining tractable.

In summary, the problem we address is to synthesize safe, real-time feasible whole-body controllers for articulated rigid-body systems by (i) globally linearizing nonlinear dynamics via Koopman operators, (ii) embedding safety constraints directly within a single MPC optimization, and (iii) adversarially adapting the safety index to ensure non-emptiness of safe control set as well as constraint solvability under learned dynamics.

%% file: sections/4_methodology.tex
\section{METHODOLOGY}

In this section we present a single linear MPC formulation for whole-body safe control of articulated robots. The approach (1) learns a globally linear lifted model via Koopman operators, (2) embeds and redesigns the safe control constraint so it remains feasible at the safe state boundary, (3) adapts the model for hardware.

\subsection{Safe MPC Formulation via Koopman Dynamics}
Although naive models such as single-integrator dynamics~\cite{ames2016control} have been widely used for safety-critical control, they are ill-suited for articulated robots: they ignore actuation limits, joint coupling, and dynamic feasibility, leading to controllers that may certify safety in theory but fail in practice on high-DOF systems. To enable tractable safe control, we  approximate \eqref{eq:manip_dyn} with a globally linear lifted model using Koopman operator. Following \eqref{eq:evolve} and \eqref{eq:evol_flow}, we define an embedding $\psi:\mathbb{R}^n \to \mathbb{R}^{n+d}$ and lifted state $z_k = [x_k; \psi_\omega(x_k)]^\top$ with $x_k = [p_k;\theta_k]^{\top}$, where $p_k \in \mathbb{R}^3$ denotes the Cartesian position of the target point on the robot for linearization (e.g., end-effector or base location for tracking, center of the collision volume for collision avoidance). The dynamics are approximated as
\begin{equation}
z_{k+1} = A z_k + B u_k, \qquad x_k = Pz_k
\label{eq:koopman_dynamics}
\end{equation}
where $u_k$ denotes joint velocity commands, $A,B$ are Koopman matrices, and $P$ projects the lifted state back to the original coordinates. By wrapping the low-level torque dynamics \eqref{eq:manip_dyn} into the Koopman approximation \eqref{eq:koopman_dynamics}, our controller operates at the velocity-control level. This design choice is crucial: it allows the use of first-order safety indices, simplifies safety constraint construction, and improves computational efficiency.
\subsubsection{Feedforward Prediction}
Given the current state $x_t$, we predict $K$ steps forward using the learned Koopman operators $A$ and $ B$, which are represented as linear layers. We roll out the dynamics:
\begin{align}
\hat{z}_{t+k+1} &= A z_{t+k} + B u_{t+k},\, k = [0, ..., K-1] \\
z_t &= \psi_x(x_t) = [x_t; \psi_\omega(x_t)]^\top \\
x_{t+k} &= P z_{t+k} \\
u_{t+k} &= \hat{u}_{t+k}
\end{align}

\subsubsection{K-steps Prediction Loss}
Since we don't have a separate network for control embedding $\psi_{u}$, we only train the state embedding network $\psi_\omega$ and Koopman matrices $A$, $B$ end-to-end using a $K$-step prediction loss. Given dataset $\{x_i, u_i\}$ for $i = 0, \ldots, K$, we compute embedded states $z_i = \psi_x(x_i)$ and predicted lifted states $\hat{z}_i$ via forward rollout. The loss is:
\begin{equation}
\mathcal{L}_{\text{pred}}(\omega) = \sum_{i=1}^{K} \gamma^{i-1} \text{MSE}(z_i, \hat{z}_i)
\end{equation}
where $\gamma$ is a decay factor and MSE is the mean squared error~\cite{shi2022deep}.

\subsubsection{Safe Linear MPC Formulation}
Using the Koopman dynamics \eqref{eq:koopman_dynamics}, the time derivative of safety constraint \eqref{eq:safety_constraint} and \eqref{eq:bphi} can be expressed as:
\begin{align}
\dot{\phi}_0 &= \frac{\partial \phi_0(\bm{x})}{\partial \bm{x}}\,\dot{\bm{x}} \nonumber\\
& \approx \frac{\partial \phi_0(\bm{x})}{\partial \bm{x}}\frac{x_{k+1} - x_k}{\Delta t}\nonumber\\
&= \nabla_x \phi_0(x_k)^\top \frac{(PA - P)z_k + PBu_k}{\Delta t}
\end{align}
This expression is linear in \(u\) and can be directly inserted as a linear constraint in MPC. Considering quadratic cost with desired state $x^{\text{des}}$, the system becomes linear in lifted space and we can formulate a quadratic programming (QP) over horizon $N$ as:
\begin{IEEEeqnarray}{rCl}
\min_{u_{0:N-1}} & \quad & \sum_{k=0}^{N-1} \lVert Pz_k - x_k^{\text{des}} \rVert_Q^2 + \lVert u_k \rVert_R^2 + \lVert Pz_N - x_N^{\text{des}} \rVert_{Q_N}^2 \nonumber \\
\text{s.t.} &       & z_{k+1} = A z_k + B u_k, \quad \forall k = 0, \dots, N-1 \nonumber \\
            &       & \nabla_x \phi_0(x_k)^\top \frac{(PA - P)z_k + PBu_k}{\Delta t} \le b_\phi(x_k), \forall k \nonumber \\
            &       & z_0 = \psi_{x}(x_0), \quad u_k \in \lbrack u_{\min}, u_{\max} \rbrack \label{eq:safe_mpc}
            \label{eq:koopman_qp}
\end{IEEEeqnarray}

\subsection{Redesign the Safety Constraint for Persistent Feasibility}
Although Koopman dynamics allows us to formulate a single QP for nonlinear systems, extra dimensions introduced by lifted space can cause feasibility issue. Let us take a more rigorous look at $\dot\phi_0$. First, matrices $A, B$ can be decomposed as:
\begin{equation}
    A = \begin{bmatrix}
        A_{xx} & A_{x\psi} \\
        A_{\psi x} & A_{\psi \psi}
    \end{bmatrix},
    B = \begin{bmatrix}
        B_x \\
        B_{\psi}
    \end{bmatrix}
\end{equation}
Given $z_k = \lbrack x_k;\psi_{\omega}(x_k) \rbrack^\top$, the lifted dynamics formulation for original state $x$ becomes:
\begin{equation}
    x_{k+1} = A_{xx}x_k + A_{x\psi}\psi_{\omega}(x_k) + B_x u_k
\end{equation}
Extra term $A_{x\psi}\psi_{\omega}(x_k)$ propagates into the time derivative computation of \eqref{eq:safety_constraint} and $\dot\phi_0=\nabla_x\phi_0^\top \dot{x} \le b_\phi$ in (\ref{eq:koopman_qp}) becomes:
\begin{IEEEeqnarray}{rCl}
&\frac{(p_{\text{ego}} - p_{\text{obs}})}{\lVert p_{\text{ego}} - p_{\text{obs}} \rVert}
\frac{\partial p_{\text{ego}}}{\partial x} 
B_x u_k
 \le  b_\phi -
\nonumber\\
&\frac{(p_{\text{ego}} - p_{\text{obs}})}{\lVert p_{\text{ego}} - p_{\text{obs}} \rVert}
\frac{\partial p_{\text{ego}}}{\partial x} (A_{xx}x_k + A_{x\psi}\psi_{\omega}(x_k) - x_k)\label{eq: constraint expanded}
\end{IEEEeqnarray}
These extra lifted contributions $A_{x\psi}\psi_{\omega}(x_k)$ can shift the half-space constraint on $u_k$ so that no bounded input $u_k\in[u_{\min},u_{\max}]$ can satisfy $\dot{\phi}_0\le b_\phi$ for all active links simultaneously. 

To adapt our safety specification to learned dynamics, we introduce Learner-Critic architecture for adversarial fine tuning. In this section, we focus on the specific setting of a 7-DoF Kinova Gen3 manipulator for clarity and conciseness. Let $d = \lVert p_{\text{ego}}(x) - p_{\text{obs}} \rVert_2$ and define a nonlinear safety index $\phi(x): \mathbb{R}^n\rightarrow \mathbb{R}$ in Cartesian coordinate as:
\begin{equation}
    \phi_{n,\beta}(x) = d_{\min}^n - d^n + \beta d
    \label{eq:safety_index}
\end{equation}
where $n, \beta \in \mathbb{R}$ are learnable parameters. This parameterization reduces the second order  formulation in \cite{liu2014control} to first order. Nevertheless, this updated formula will reshape the gradients in \eqref{eq: constraint expanded} to make it easier to find the corresponding  $u_k$ within the control limit and potentially conflicting constraints from multiple collision volumes.  Let $\Gamma \doteq (n,\beta)$ and learning is formulated as a min-max optimization problem:
\begin{equation}
    \min_{\Gamma} \max_{x \in \partial{S}} \mathcal{L}_{\text{infeas}}(\Gamma,x)    
\end{equation}
where the \textit{infeasibility risk} is: 
\begin{equation}
    \mathcal{L}_{\text{infeas}}(\Gamma,x) \coloneqq \min_{v \in \mathcal{V}(\mathcal{U})} \dot\phi_{\Gamma}(x, v)
\end{equation}

Unlike the original formulation in \cite{pmlr-v205-liu23e}, where the safety index $\phi(x)$ is scalar and defined over the entire system, the application on the 7-DOF Kinova arm requires each link to be evaluated for safety independently. We define a linkwise safety index $\phi^i(x)$ for each link $i = 1,\dots,7$, and consider all of them as safety constraints. This linkwise definition introduces substantial complexity into both the constraint evaluation and the process of counterexample collection, compared to \cite{pmlr-v205-liu23e}, where only a single constraint is considered.

To enable adversarial fine tuning under this setup, we modify a Learner–Critic architecture. The Critic attempts to collect boundary states and associated control counterexamples that violate the constraint $\dot{\phi}^i(x, u) \le 0$ for all active links at the safety boundary. However, instead of performing explicit Projected Gradient Descent (PGD) on \textit{infeasibility risk} \cite{madry2017towards}, our implementation uses a two-stage process that achieves an equivalent effect, adapted to the multi-link robotic setting.

First, boundary states are sampled using a differentiable optimization procedure. Specifically, in the presence of an obstacle in 3D space we minimize the distance between any of the robot link's center-of-mass positions and the obstacle center: 
\begin{equation}
\mathcal{L}_{\text{dist}}(x) = \sum_{i=1}^7 \max \{0,\, d_{\min} - \lVert p_i(x) - p_{\text{obs}} \rVert_2\}
\end{equation}
{\color{black}
We emphasize that this projection step is defined using the \emph{geometric} safety specification $d=d_{\min}$ (a $0.2$\,m clearance shell) rather than the learned level set $\{\phi_{n,\beta}=0\}$ as done in \cite{pmlr-v205-liu23e}. Anchoring counterexample collection to the physical clearance requirement ensures that the Critic explicitly probes where infeasibility first emerges for the original safety margin, while $\phi_{n,\beta}$ is subsequently tuned to restore feasibility under Koopman dynamics. This is also empirically plausible as we are working in a MPC setting instead reactive safety filtering setting. }
Gradient descent is used to optimize $x$ until this projection loss reaches a small threshold (empirically set to $0.0001$), ensuring that the robot is brought close to the safety boundary. This serves as a practical projection step, and thus effectively making this process a PGD.

Next, for each near-boundary state, the Critic samples control candidates from a set of saturated control vertices $\mathcal{V}(\mathcal{U})$. For each control $u$, it computes \textit{infeasibility risk}. A counterexample is recorded only if no control in the admissible set satisfies:
\begin{equation}
    \mathcal{L}_{\text{infeas}}^i(\Gamma,x) \le 0, \quad\forall i \in \partial{\mathcal{S}}
\end{equation}
for all links $i$ in contact with the boundary.

To prevent the safety index from becoming overly conservative, which could increase infeasibility, we bound the Critic’s influence by allowing 10 trials to collect the desired number of counterexamples (e.g., 50). If this quota is not reached, we consider it sufficiently tuned. This guards against overfitting to pathological states, which would otherwise cause the Learner to reshape $\phi$ in an attempt to satisfy constraints that cannot be satisfied, leading to an overly conservative and distorted safety index.

\begin{table}[t]
\centering
\footnotesize
\setlength{\tabcolsep}{6pt}
\renewcommand{\arraystretch}{1.15}
\begin{tabular}{l c}
\hline
\textbf{Scenario} & \textbf{\# Infeasible / $T_{\mathrm{exp}}$} \\
\hline
Single Obstacle (Untrained) & $108/4000$ \\
Multi Obstacle (Untrained)  & $632/4000$ \\
Single Obstacle (\textbf{Trained})   & $42/4000$ \\
Multi Obstacle (\textbf{Trained})    & $113/4000$ \\
\hline
\end{tabular}
\caption{QP infeasibility counts comparison. Single Obstacle includes one dynamic obstacle chasing one of Kinova's links, whereas Multi Obstacle includes seven extra static obstacles placed at random positions. This experiment is conducted without slack relaxation. Each entry reports the number of infeasible QP solves among the total control steps in the experiment ($T_{\mathrm{exp}}=4000$).}
\label{tab:qp_infeasibility}
\end{table}

The Learner then performs a gradient update on the safety index parameters $\Gamma$ by minimizing a combined loss:
\begin{equation}
\mathcal{L}_{\text{total}} = \frac{1}{M} \sum_{x \in \mathcal{B}} 
\mathbb{E}_{i \in \partial\mathcal{S}} \Big\lbrack \mathcal{L}_{\text{infeas}}^i(\Gamma,x) \Big\rbrack 
+ \mathcal{L}_{\text{reg}}
\label{eq:total_loss}
\end{equation}
where $\mathcal{B}$ is the batch of counterexamples, $M$ is the size of a batch, $\mathcal{L}_{\text{reg}} = \mu \lVert \Gamma-\Gamma_0 \rVert_2^2$ is a regularization loss  penalizing deviation from heuristically chosen good initial values $\Gamma_0=(n_0, \beta_0)$. The overall procedure is summarized in Algorithm~\ref{euclid}.

The final MPC problem then becomes:
\begin{IEEEeqnarray}{rCl}
\min_{u_{0:N-1}} & \quad & \sum_{k=0}^{N-1} \lVert Pz_k - x_k^{\text{des}} \rVert_Q^2 + \lVert u_k \rVert_R^2 + \lVert Pz_N - x_N^{\text{des}} \rVert_{Q_N}^2 \nonumber \\
\text{s.t.} &       & z_{k+1} = A z_k + B u_k, \quad \forall k = 0, \dots, N-1 \nonumber \\
            &       & \nabla_x \phi_{n,\beta}^i(x_k)^\top \frac{(PA - P)z_k + PBu_k}{\Delta t} \le b_\phi(x_k),\forall i,k \nonumber \\
            &       & z_0 = \psi_{x}(x_0), \quad u_k \in \lbrack u_{\min}, u_{\max} \rbrack \label{eq:safe_mpc_new}
            \label{eq:koopman_qp_new}
\end{IEEEeqnarray}

The effectiveness of adversarial fine tuning in reducing infeasible cases is quantified in Table~\ref{tab:qp_infeasibility}, which compares QP infeasibility counts across different obstacle settings before and after training.

\begin{algorithm}[t]
\caption{Adversarial Fine Tuning}\label{euclid}
\small
\begin{algorithmic}[1]

\Require{$\Gamma_0=(n_0,\beta_0)$, $p_{\text{obs}}$, $d_{\min}$ regularization weight $\mu$, batch size $N_{\text{batch}}$, trials for Critic $N_{\text{trials}}$}

\Function{CollectCE}{$\Gamma$, $p_{\text{obs}}$, $d_{\min}$, $N_{\text{batch}}$, $N_{\text{trials}}$}
    \State Set learning rate $\alpha_{c}$, $N_{\text{collected}} = 0$
    \For{$N_{\text{trials}}$}
    \State $X \leftarrow$ uniformly sample within operating region
    \State $X \leftarrow X + \alpha_c\cdot\nabla_X\mathcal{L}_{\text{dist}}$
    \State $X_{\text{CE}}\leftarrow\min_{v\in\mathcal{V}(\mathcal{U})}\mathcal{L}_i>0 \forall i=1,...,7$
    \If{$N_{\text{collected}} \ge N_{\text{batch}}$}
        \State\Return $(X_{\text{CE}}, U_{\text{CE}})$
    \EndIf
    \EndFor
\State \textit{Training complete}
\EndFunction
\Function{Update}{$\Gamma$, $X_{\text{CE}}$, $U_{\text{CE}}$}
    \State Set learning rate $\alpha_l$ 
    \State $\Gamma \leftarrow \Gamma + \alpha_l\cdot\nabla_{\Gamma}\mathcal{L}_{\text{total}}$ \hfill \text{(from Eq. \eqref{eq:total_loss})}
    \State\Return $\Gamma$
\EndFunction

\Function{Main}{}
    \State $\Gamma=(n_0,\beta_0)$
    \While{not done}
    \State $(X_{\text{CE}}, U_{\text{CE}})\leftarrow$ CollectCE($n,\beta$)
    \State $\Gamma\leftarrow$ Learn($\Gamma,X_{\text{CE}},U_{\text{CE}}$)
    \EndWhile
    
    \Return $\Gamma$
\EndFunction
\end{algorithmic}
\end{algorithm}


\subsection{Sim to Real Adaptation}
To bridge the gap between simulation and hardware, we first quantify the sim-to-real mismatch by executing identical reference trajectory tracking experiments in PyBullet simulation and on the real robot. The deviation in tracking performance serves as a measure of model discrepancy. Instead of retraining the entire Koopman embedding and operator matrices, we collect hardware data and fine-tune only the $A$ and $B$ matrices of the lifted linear dynamics \eqref{eq:koopman_dynamics}. This lightweight adaptation step efficiently accounts for actuation and unmodeled dynamics differences between simulation and hardware, enabling migration with minimal retraining. As a result, the controller preserves the structure of the learned embedding while maintaining high accuracy in real-world deployment.

\begin{figure}[t]
    \centering
    \includegraphics[width=0.5\columnwidth]{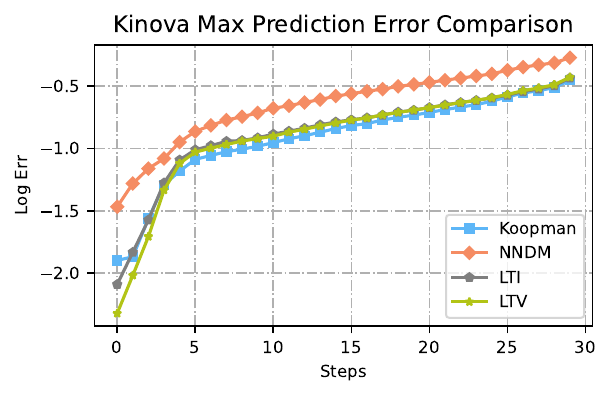}
    \hspace{-3.5mm} 
    \includegraphics[width=0.5\columnwidth]{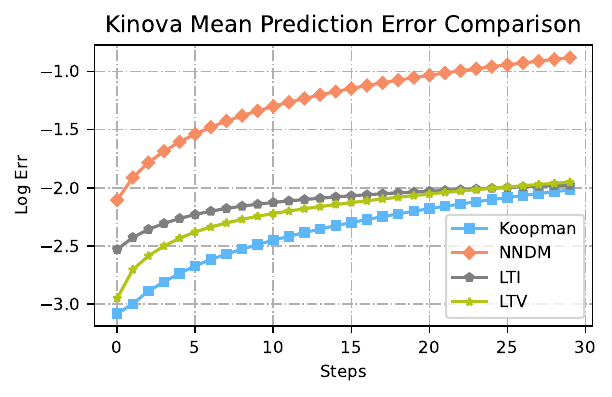}
    \caption{Prediction Error Comparison. The analytic baselines (LTI and LTV), KDM, and NNDM are compared against PyBullet ground truth. KDM maintains the lowest long-horizon error growth, while the analytic model is competitive in short horizons but diverges.}
    \label{fig:cmp_err}
\end{figure}

%% file: sections/5_experiments.tex
\section{EXPERIMENTS}

\begin{table*}[t]
\centering
\footnotesize
\setlength{\tabcolsep}{3pt}
\renewcommand{\arraystretch}{1.15}
\resizebox{0.95\textwidth}{!}{%
\begin{tabular}{|l||
                c|
                c|
                c|c|c|c|c|}
\hline
\multirow{2}{*}{} &
\multicolumn{1}{c|}{\textbf{Single Obstacle}} &
\multicolumn{1}{c|}{\textbf{Multi Obstacle (6 total)}} &
\multirow{2}{*}{\centering\arraybackslash\makecell{\textbf{Avg. Distance}\\\textbf{to Target [m]}}} &
\multirow{2}{*}{\centering\arraybackslash\makecell{\textbf{Avg. Max $\phi$}\\\textbf{Over Links}}} &
\multirow{2}{*}{\centering\arraybackslash\makecell{\textbf{Avg. Mean $\phi$}\\\textbf{Over Links}}} &
\multirow{2}{*}{\centering\arraybackslash\makecell{\textbf{Avg. Min Dist.}\\\textbf{to Obstacle [m]}}} &
\multirow{2}{*}{\centering\arraybackslash\makecell{\textbf{Cumulative}\\\textbf{Cost}}} \\ \cline{2-3}
 & \makecell[c]{\centering\textbf{Avg. Comp. Time [s]}} &
   \makecell[c]{\centering\textbf{Avg. Comp. Time [s]}} &
   & & & & \\ \hline
\textbf{KMPC (Ours)} & 9.66e-3 $\pm$ 0.15e-3 & 1.496e-2 $\pm$ 0.23e-3 & 0.071860 & -0.03828 & -0.21492 & 0.21913 & 13154 \\ \hline
LTIMPC                & 5.19e-3 $\pm$ 0.43e-3 & 0.629e-2 $\pm$ 0.79e-3 & 0.572857 & -0.15237 & -0.33137  & 0.32237 & 307378578 \\ \hline
LTVMPC                & 14.98e-3 $\pm$ 0.12e-3 & 0.629e-2 $\pm$ 0.79e-3 & 0.144116 & -0.09723 & -0.18294  & 0.26745 & 93292 \\ \hline
NMPC-10             & 40.49e-3 $\pm$ 0.96e-3 & 5.299e-2 $\pm$ 3.06e-3 & 0.116499 & -0.16023  & -0.25239 & -      & 81261 \\ \hline
NMPC-100            & 412.59e-3 $\pm$ 227.05e-3 & 1.04704 $\pm$ 0.03220 & 0.009787 & -0.12234 & -0.25224 & - & 160 \\ \hline
\end{tabular}%
}
\caption{Computation Time and Performance Comparison for an episode length of 4000 control steps. 
The two left columns report average computation time for safe control, while the remaining columns summarize tracking performance and safety in the 3D space.
Cumulative cost reflects tracking performance.
For safety, \textbf{Avg. Max $\phi$} denotes the maximum $\phi$ over all links at each time step, averaged over the episode, and \textbf{Avg. Mean $\phi$} denotes the mean $\phi$ over links at each time step, averaged over the episode; in both cases, lower values indicate safer behavior. \textbf{Avg. Min Dist. to Obstacle} reports the minimum link-to-obstacle distance averaged over the episode, where larger values indicate safer separation.
The symbol “--” indicates that the robot collided with the obstacle during the experiment.
Across these complementary metrics, KMPC achieves the \emph{best performance--safety trade-off}, attaining low tracking cost while making only necessary evasion, thereby maintaining moderately low $\phi$ (both Avg. Max and Avg. Mean).}

\label{tab:time_perf}
\end{table*}

In this section, we conduct experiments using a Kinova Gen3 manipulator and a Unitree Go2 to evaluate our approach. 

\subsection{Experiment Setup}
For all dynamics models considered—Koopman Dynamics Model (KDM), Neural Network Dynamics Model (NNDM) \cite{wei2022safe}, 
and the analytic baselines—the state is defined as the concatenation of target link position $p_k$ and joint angles $q_k$: $x_k = \begin{bmatrix} p_k ;q_k \end{bmatrix}^\top$. 

\subsubsection{Analytic Baselines}
We compare two analytic formulations: a time-invariant (LTI) model and a time-varying (LTV) variant.
Both share the same state update for the joint angles and end-effector:
\begin{align}
    p_{k+1} &= p_k + \Delta t \, J(\bar q_k) \, u_k \\
    q_{k+1} &= q_k + \Delta t \, u_k
\end{align}
where $u_k$ is the joint velocity command and $J(\bar q_k)$ denotes the manipulator Jacobian evaluated at 
a reference configuration $\bar q_k$.
For the LTI baseline, $\bar q_k = q^\ast$ is fixed, giving a globally linear but coarse approximation.
For the LTV system, $\bar q_k = q_k$ is updated at each step, yielding a locally valid linearization that 
improves tracking while keeping $q_k$ integrated directly.  
This LTV formulation supports an MPC controller (LTVMPC) that bridges the gap between a static linear model 
and full nonlinear MPC.

\subsubsection{Learned Models}
We train NNDM and KDM within PyBullet simulation. Both employ fully connected feedforward networks with 
hidden layers [256, 256, 256] for either the lifting function $\psi_\omega$ (KDM) or direct dynamics mapping (NNDM). 
Fig. \ref{fig:cmp_err} compares the prediction errors across models, showing that the Koopman-based network achieves the lowest prediction error.
Once trained, we formulate linear MPC for KDM (KMPC), LTI, and LTV baselines, and shooting-based NMPC for the other models. 
For all non-Koopman models, a safety filter is additionally applied to enforce state constraints.

\subsubsection{Controller and solver setup}
All QP-based controllers (KMPC, LTIMPC, LTVMPC, and the safety filters) are solved using \texttt{OSQP}. For a fair comparison, we use the same MPC horizon $H=9$ for all methods with control input a 7D joint velocity command $u_k \in \mathbb{R}^7$ along with box limits matching the Kinova joint velocity saturation. We use a first-order safety bound with $\lambda=0.05$ in all rollouts. For numerical stability, slack relaxation is used whenever we enforce safety constraints via a QP: KMPC includes slack directly in its unified QP, while the analytic baselines and NNDM apply slack in the separate safety-filter QP.

\begin{figure}[t]
    \centering
    \includegraphics[width=\columnwidth]{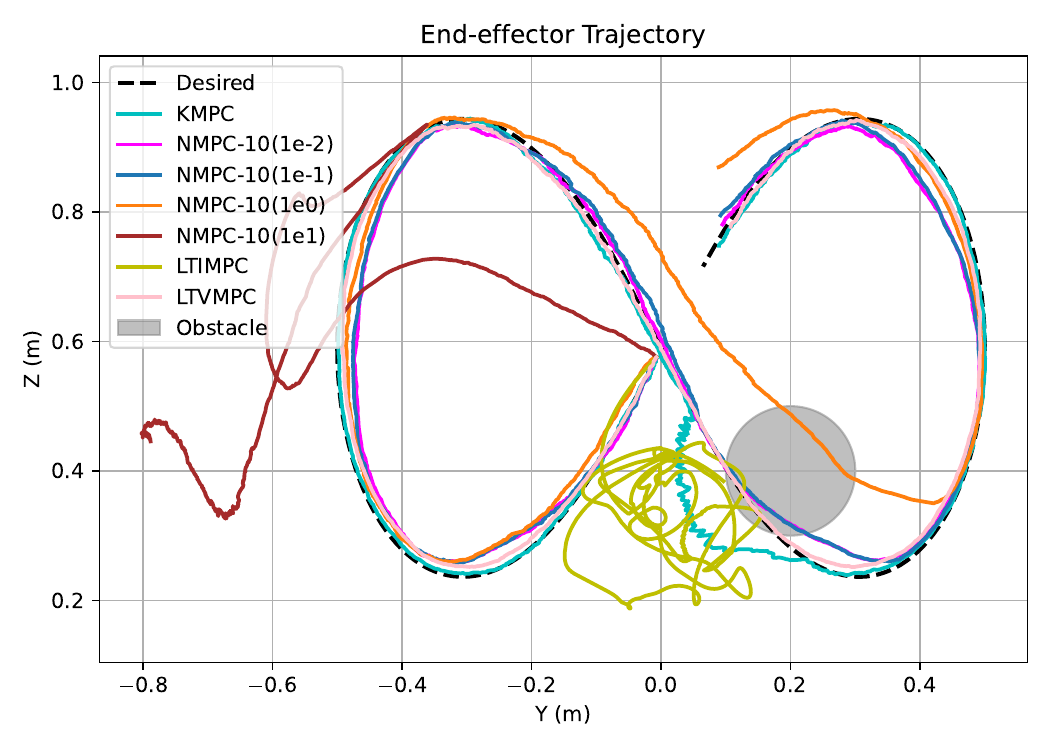}
    \caption{\footnotesize Safe Tracking Comparison. NMPC-10 is nonlinear MPC with 10 times shooting, values in parentheses represent slack weight for safety filter relaxation. Linear counterparts, including KMPC did not require slack relaxation for this experiment. KMPC outperforms NMPC-10 both in tracking and safety constraint satisfaction.}
    \label{fig:twofigs}
\end{figure}

\begin{figure*}[t]
    \centering
    \includegraphics[width=0.85\textwidth]{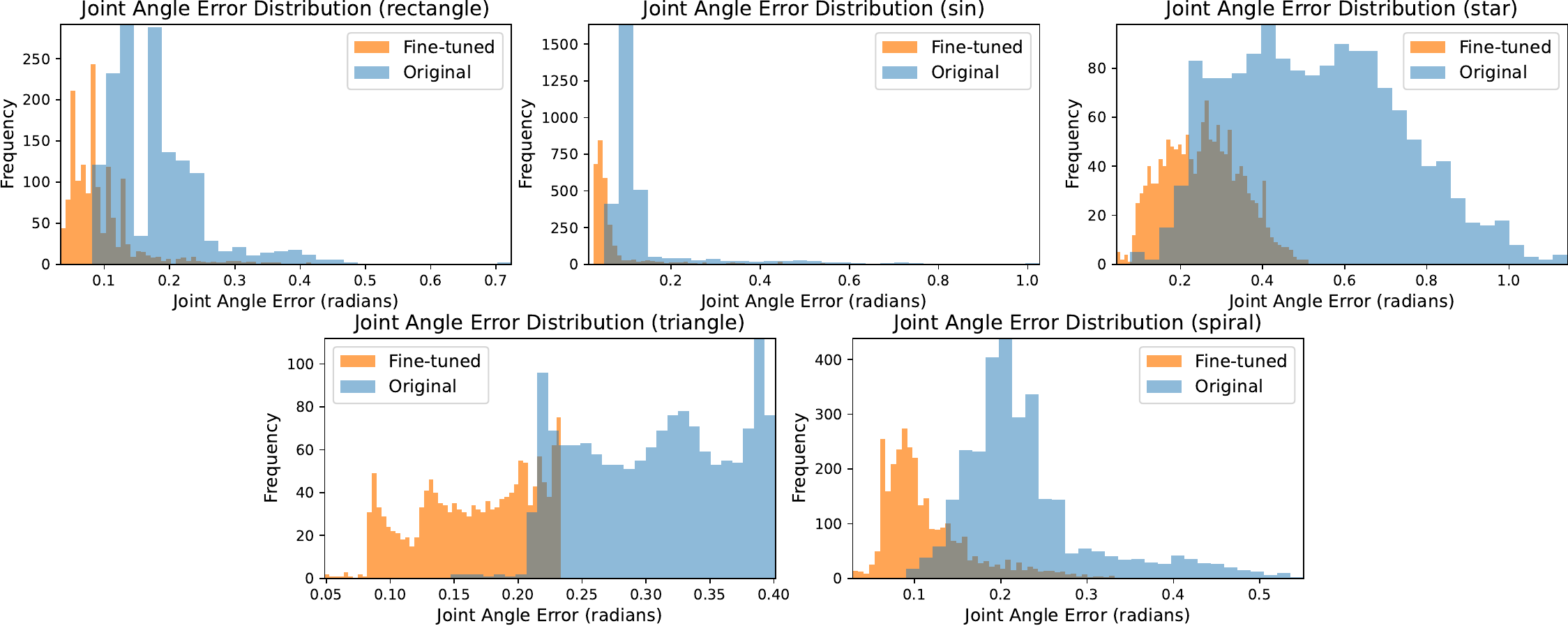}
    \caption{Histogram of single-step joint-angle prediction error norms (radians) over five hardware trajectories (rectangle, sin, star, triangle, spiral), comparing the original Koopman model (blue) against the fine-tuned Koopman model (orange). Fine-tuning shifts the error distributions toward smaller values across all trajectories, yielding a mean joint-angle error of 0.140 rad (maximum 0.546 rad) and a mean end-effector position error of 0.031 m (maximum 0.143 m) on the evaluated hardware dataset.}
    \label{fig:sim_to_real_joint}
\end{figure*}

\begin{figure}[htbp!]
    \centering
    \includegraphics[width=1.0\columnwidth]{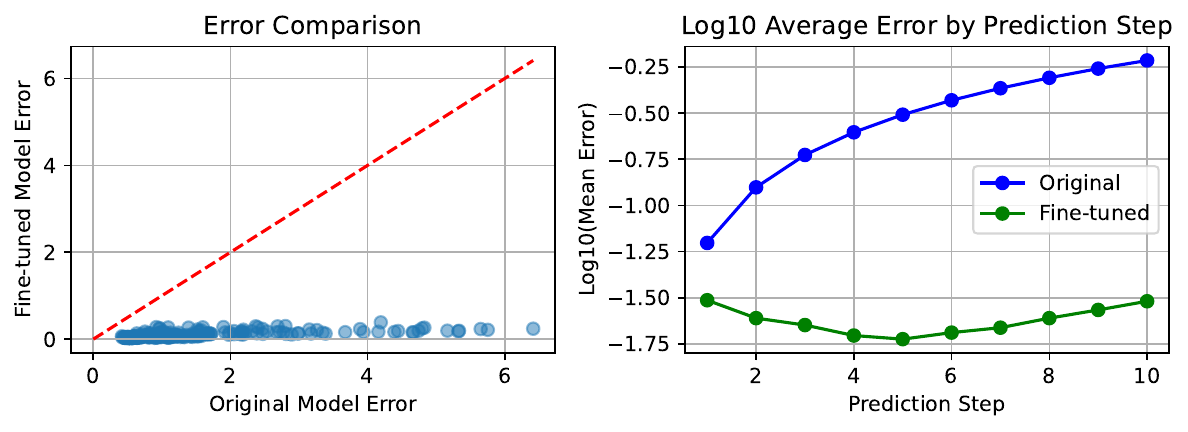}
    \caption{Prediction error comparison for Kinova-Gen3 before and after fine-tuning. 
(Left) Scatter plot of original versus fine-tuned model errors, where nearly all points lie below the $y=x$ line (red), indicating consistent improvement; (Right) Average prediction error across horizons (log$_{10}$ scale), showing that fine-tuned Koopman dynamics maintain significantly lower error over multiple steps. }
    \label{fig:sim_to_real}
\end{figure}

\subsection{Safe Control in 2D Space}
This experiment incorporates collision avoidance of the \textit{end-effector} as a safety constraint. We use the first-order safety index from \eqref{eq:safety_spec} with $d_{\min} = 0.2$ for NNDM and the analytic baseline, and the adapted safety index from \eqref{eq:safety_index} for KDM. The KDM control law solves the QP in \eqref{eq:koopman_qp}, whereas MPC for the other models solve equivalent tracking problem in original state space with the safety filter formulated as:
\begin{IEEEeqnarray}{rCl}
\min_{u_k} & \quad & \lVert u_k - u_k^{\text{ref}} \rVert^2 \nonumber \\
\text{s.t.} &       & \nabla_x \phi_0(x_k)^\top \frac{x_{k+1}-x_k}{\Delta t} \le b_{\phi}(x) \label{eq:qp}
\end{IEEEeqnarray}

We employ a 9-step horizon for all models to ensure consistent comparison. Results depicted in Fig.~\ref{fig:twofigs} demonstrate effective obstacle avoidance and accurate tracking of KMPC.

\subsection{Safe Control in 3D Space}
In this scenario, the robot is tasked with tracking a reference end-effector trajectory in full 3D space while avoiding static obstacles placed along the path. This setting requires simultaneous trajectory tracking and obstacle avoidance across all seven links. The MPC optimization problem is identical except for the linkwise-extended safety constraint:
\begin{align}
\dot{\phi}^i(x,u) \le b_{\phi}(x),
            \quad \forall i=1,...,7
\end{align}

The performance results in Table~\ref{tab:time_perf} illustrate robust trajectory tracking with obstacle evasion, validating the practicality of KMPC frameworks for safe robot control. While KMPC does not attain the numerically best value for every individual safety metric ($\phi$) or cumulative cost, the aggregate results indicate that KMPC selects the most favorable operating point between aggressive tracking and conservative safety among all baselines. LTI and LTV baselines show poor tracking cost (over $7.1\times$ higher than KMPC) due to model mismatch and increased penalty from constraint handling, while NMPCs risks colliding into the obstacle. The proposed KMPC also achieves faster computation compared to NMPC (over $4.2\times$ faster) by exploiting linear Koopman dynamics, whereas shooting-based MPC becomes computationally impractical in these scenarios.

\subsection{Sim-to-Real Adaptation}

To evaluate the robustness of our approach beyond simulation, we deployed the fine-tuned Koopman-based safe control pipeline directly on the Kinova Gen3 manipulator. Figure~\ref{fig:sim_to_real} compares state prediction errors before and after fine-tuning, showing substantial reduction in both joint angle and end-effector position errors. These results confirm that the proposed methodology not only improves model accuracy in simulation but also transfers effectively to real hardware without retraining the embedding. Our experiments demonstrate that the unified pipeline, combining Koopman lifting and adversarial safety index tuning, achieves reliable performance on the physical system. A supplementary demo video is provided, illustrating safe control execution on hardware and validating the practical deployability of the proposed framework.

Figure~\ref{fig:sim_to_real_joint} shows that fine-tuning consistently shifts the joint-angle error histograms left across all five hardware trajectories, reducing both the typical error and the probability of large-error events (i.e., a thinner right tail). 
Importantly, this improvement is achieved by adapting only the linear Koopman operators while keeping the lifting/embedding fixed, suggesting that the learned latent coordinates remain transferable and that the dominant sim-to-real mismatch is captured by linear dynamics in lifted space. 
Although the fine-tuned model reduces the mean error to 0.140~rad, the remaining worst-case errors (up to 0.546~rad) indicate occasional transients likely caused by unmodeled hardware effects such as actuation delays, friction, and state-estimation noise. 
The corresponding reduction in end-effector error (mean 0.031~m, max 0.143~m) confirms that the improvement is not limited to configuration space but also translates to task-space accuracy. 

Collectively, these experiments demonstrate the versatility and effectiveness of our proposed Koopman-based control methodology, from simple trajectory tracking to dynamic safety-critical scenarios. It is worth noting that since safety specification is derived from Koopman space, successful evasion also depicts the accuracy in model approximation. 

\subsection{Extension to Floating-Base Systems}
Unlike fixed-base manipulators, floating-base robots introduce additional unactuated base degrees of freedom whose motion is coupled with the actuated joints, so the state must explicitly include the base pose/velocity and safety constraints must be enforced over the full-body (base + joints) kinematics. readers can refer to the Appendix for the complete formulation details and additional Unitree Go2 results.

%% file: sections/6_future_work.tex
\section{FUTURE WORK}
While the proposed Koopman-based control framework shows promising results, several limitations remain and motivate future directions. Our current formulation relies on a first-order safety index based on positional information, which may be insufficient for dynamic or fast-evolving environments. Future work will explore higher-order safety indices that incorporate velocity and curvature, likely requiring torque- or acceleration-level control. These extensions will further improve the scalability, safety, and real-world viability of Koopman-based safe control.

%% file: sections/7_conclusion.tex
\section{CONCLUSION}
This project presents a data-driven safe control framework that combines Koopman operator theory, Safe Set Algorithm, and adversarial fine tuning to enable efficient, verifiable control of nonlinear robotic systems. By lifting nonlinear dynamics into a latent linear space via a neural embedding, we exploit classical linear control techniques while enforcing safety constraints through a single QP.
We validate the approach on the Kinova Gen3 manipulator and Unitree Go2 in tasks ranging from trajectory tracking to dynamic obstacle avoidance with multi-step MPC, demonstrating reliable tracking and real-time safety. The framework’s linear structure also facilitates adaptive extensions and hardware deployment. Looking forward, we aim to scale to higher-dimensional systems and incorporate higher-order safety indices.
Overall, Koopman-based safe control offers a scalable and interpretable alternative to conventional model-free methods, particularly in safety-critical robotics.

%% file: sections/8_appendix.tex


\subsection*{A. Extension to Floating-Base Systems}
\begin{figure}[t]
    \centering
    \includegraphics[width=1.0\columnwidth]{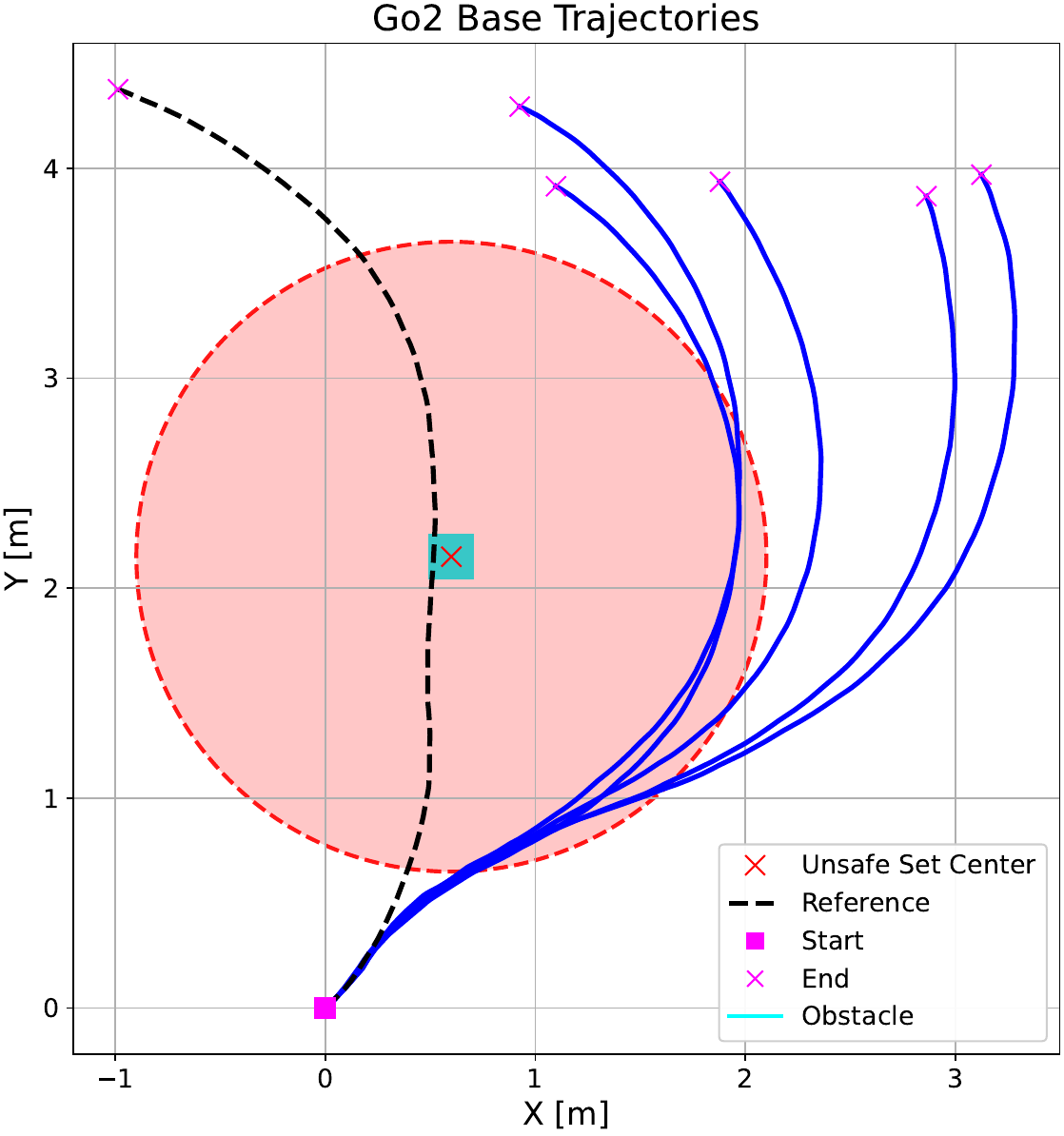}
    \caption{Go2 base trajectories over five runs. The reference (black) passes through the unsafe region (red), whereas Koopman Safe MPC produces evasive behavior and steers the robot back toward safety. Run-to-run variation is primarily due to omitting globally-referenced base translation $(x,y)$ from the predictive state, so the controller does not explicitly regulate absolute global position.}
    \label{fig:go2_plot_2d}
\end{figure}

\begin{figure}[h]
    \centering
    \includegraphics[width=1.0\columnwidth]{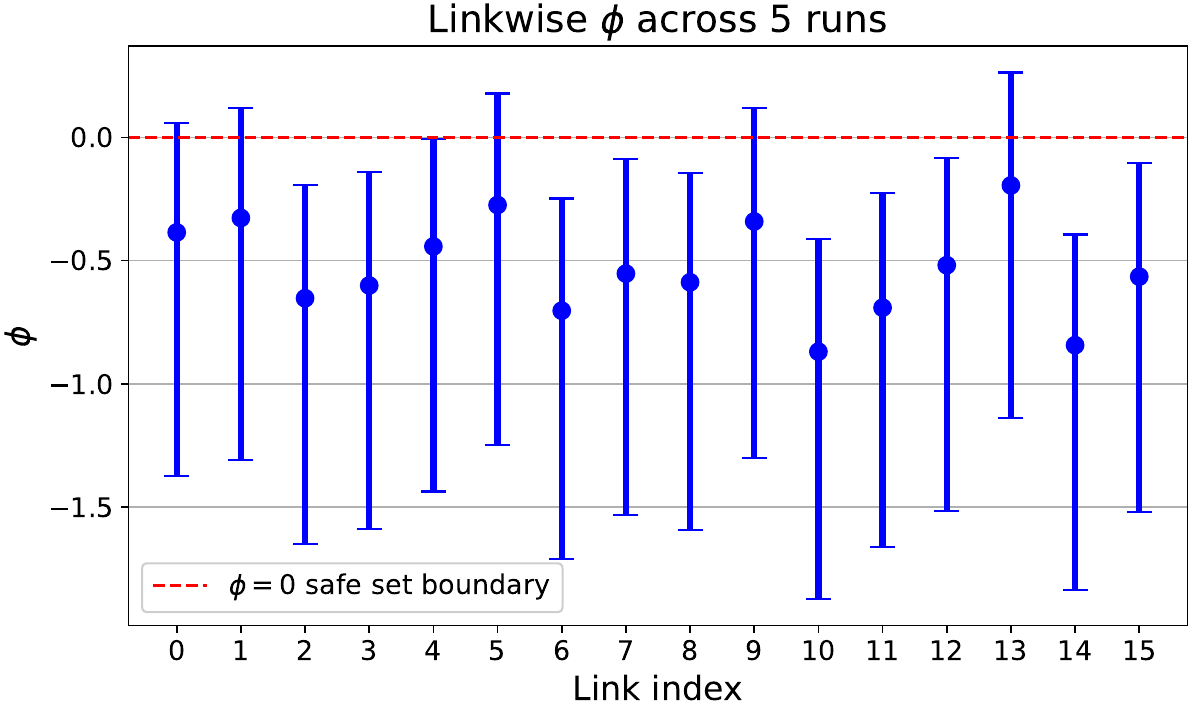}
    \caption{Linkwise safety index $\phi$ across five runs. Dots indicate the mean linkwise $\phi$ per link, with whiskers showing min/max across runs. Positive values correspond to temporary unsafe-set violations induced by the reference; nevertheless, the overall trend shows rapid recovery back toward $\phi \le 0$ after avoidance maneuvers.}
    \label{fig:go2_linkwise_phi}
\end{figure}

While the main body of this paper focuses on fixed-base manipulation, we additionally evaluate the proposed Koopman Safe MPC on a floating-base legged system (Unitree Go2) in Isaac simulation (Fig.~\ref{fig:go2_plot_2d}--\ref{fig:go2_linkwise_phi}). This appendix clarifies the key modeling and safety-constraint differences between fixed-base and floating-base robots, and explains the practical controller implementation used in our Go2 experiments.

\paragraph{Difference between fixed-base and floating-base systems for safe control.}
For fixed-base manipulators, the controlled configuration fully determines the world pose of each link through forward kinematics. Consequently, the signed-distance safety specification (\ref{eq:safety_spec}) is naturally expressed as a function of joint configuration $q$:
\begin{equation}
    \phi_0(q) = d_{\text{min}} - \lVert p_{\text{ego}}(q) - p_{\text{obs}}  \rVert_2
    \label{eq:old_safety_spec}
\end{equation}
In contrast, for floating-base robots, link positions in the world depend on both joint configuration and the base pose. In practice, relying on globally-referenced base translation (e.g., absolute $x,y$ in a world frame) can be brittle in real environments due to drift and intermittent global localization.
Therefore, in our Go2 setup we construct the state using joint positions/velocities and base orientation/velocities, without the global base translation components ($x,y$). This choice is consistent with legged locomotion policies that are typically trained and executed in a locally-referenced representation.

\paragraph{Whole-body kinematics and safety derivative.}
The essential difference in safety constraint construction is the kinematic mapping from generalized motion to link Cartesian velocity.
For fixed-base manipulators, the link velocity depends only on joint motion:
\begin{equation}
v_{\text{link}}^w = J_q(q)\,\dot q
\label{eq:v_link_fixed}
\end{equation}
where $J_q$ denotes a link position's Jacobian with respect to configuration, and $\dot\phi$ can be written using the joint Jacobian alone.

\smallskip
Throughout the paper, we write the safety derivative as $\dot\phi=\nabla_x\phi(x)^\top \dot x$.
For a distance-based safety specification $\phi(x)=\varphi(d(x))$ with $d(x)=\|p_{\text{obs}}^w-p_{\text{link}}^w(x)\|_2$, the chain rule gives
\begin{equation}
\dot\phi
= \nabla_x\phi(x)^\top \dot x
= \varphi'(d)\,\dot d
= \varphi'(d)\,\hat u_d^\top\bigl(v_{\text{link}}^w - v_{\text{obs}}^w\bigr)
\label{eq:chain_rule_equiv}
\end{equation}
where $\hat u_d := (p_{\text{obs}}^w - p_{\text{link}}^w)/\|p_{\text{obs}}^w - p_{\text{link}}^w\|_2$ is the unit vector pointing from the link toward the obstacle center and $\varphi'(d)$ is a scalar factor (e.g., $\varphi'(d)=-1$ for signed distance; for adversarially fine-tuned indices such as $d_{\min}^n-d^n-\beta d$, $\varphi'(d)=-(n d^{n-1}+\beta)$).
Thus, the kinematic expression is not a different definition, it is a specialization of $\nabla_x\phi^\top \dot x$ for distance-based $\phi$.

\smallskip
In the fixed-base setting, we evaluate $\dot x$ using the learned Koopman rollout:
\begin{equation}
\dot x_k \approx \frac{x_{k+1}-x_k}{\Delta t},\quad
x_{k+1}=P(Az_k+Bu_k)
\end{equation}
which yields a control-affine form in $u_k$ and enables enforcing $\dot\phi$ constraints at each step of the MPC horizon using the predicted state sequence.

\smallskip
However, for floating-base systems, the link world velocity includes both base and joint contributions:
\begin{equation}
v_{\text{link}}^w = J_b(x)\,v_{\text{base}} + J_q(x)\,\dot q
\label{eq:whole_body_vel}
\end{equation}
where $v_{\text{base}}$ denotes the base spatial velocity (twist) and $J_b(\cdot)$, $J_q(\cdot)$ are the corresponding whole-body Jacobian blocks.
For the signed distance $\phi_0 = d_{\min}-\|p_{\text{obs}}^w-p_{\text{link}}^w\|_2$, the time derivative satisfies
\begin{equation}
\dot\phi_0 = \hat u_d^\top\bigl(v_{\text{link}}^w - v_{\text{obs}}^w\bigr)
\end{equation}
which reduces to $\hat u_d^\top v_{\text{link}}^w$ for static obstacles.
Substituting~\eqref{eq:whole_body_vel} yields a control-affine inequality in $\dot q$:
\begin{equation}
\underbrace{(\hat u_d^\top J_q)}_{L_g\phi}\dot q
\;\le\;
b_\phi(x)
\;-\;
\underbrace{\hat u_d^\top(J_b v_{\text{base}}-v_{\text{obs}}^w)}_{L_f\phi}
\label{eq:go2_lf_lg}
\end{equation}
where the base-motion term acts as a known drift (measured from the robot state at the current time), and the decision variable remains the joint-velocity command $u_k = \dot q_k$.
This is the key modification from the fixed-base case: ignoring $J_b v_{\text{base}}$ can severely mis-estimate $\dot\phi$ and cause safety constraints to activate too late or appear infeasible.

\smallskip
In summary, equations~\eqref{eq:chain_rule_equiv}--\eqref{eq:go2_lf_lg} unify the two viewpoints used in this paper.
For fixed-base manipulation, we compute $\dot\phi$ along the MPC horizon by combining (i) the chain rule $\dot\phi=\nabla_x\phi^\top\dot x$ with (ii) Koopman-based finite-difference prediction $\dot x\approx(x_{k+1}-x_k)/\Delta t$.
For floating-base locomotion, since our predictive state omits global base translation, we compute $\dot\phi$ at the velocity level using whole-body kinematics, i.e., (\ref{eq:whole_body_vel}).
This preserves control-affinity in the commanded joint velocities while explicitly capturing base-induced link motion.
Extending this construction to time-varying horizon-wise geometry (i.e., updating $\hat u_d$, $J_b$, and $J_q$ at each predicted step) would require additional global pose propagation or surrogate kinematics rollouts, and is left for future work.

\paragraph{KMPC formulation.}
In our Go2 experiment, we embed the safety constraint into the same QP used for tracking, consistent with the unified Koopman Safe MPC philosophy.
However, enforcing~\eqref{eq:go2_lf_lg} with exact per-step geometry across the full horizon requires evaluating $\hat u_d$, $J_b$, and $J_q$ at each predicted state.
Because these quantities depend on forward kinematics and normalization in the world frame, they are nonlinear functions of the predicted configuration.
In our current implementation, we adopt a standard linear-MPC approximation: we evaluate $(u_d, J_b, J_q)$ at the current measured state and reuse this linearization along the horizon.
\begin{IEEEeqnarray}{rCl}
\min_{u_k} & \quad & \sum_{k=1}^{N-1} \lVert Pz_k - x_k^{\text{des}} \rVert_Q^2 + \lVert u_k \rVert_R^2 + \lVert Pz_N - x_N^{\text{des}} \rVert_{Q_N}^2 \nonumber \\
\text{s.t.} &       & z_{k+1} = A z_k + B u_k, \quad \forall k = 0, \dots, N-1 \nonumber \\
            &       & \bigl(\hat u_d^\top J_q(x_t)\bigr)\,u_k \le b_\phi(x_t) \nonumber \\
            &       & \quad -\ \hat u_d^\top\!\Bigl(J_b(x_t)\,v_{\text{base},t}-v_{\text{obs},t}^w\Bigr), \quad \forall k=0,\dots,N-1 \nonumber \\
            &       & z_1 = \psi_{x}(x_1), \quad u_k \in \lbrack u_{\min}, u_{\max} \rbrack 
\label{eq:kmpc_go2}
\end{IEEEeqnarray}
This yields a tractable QP while capturing the dominant floating-base drift term through $v_{\text{base}}$.
Developing a fully time-varying horizon linearization and incorporating robust global pose propagation is left as future work.

\paragraph{Koopman model and controller setup.}
We use a Koopman dynamics model with a ResNet-3 encoder $\psi_\omega$ with hidden dimensions [256, 256, 256] to lift the state into a linear latent space. The lifted dynamics evolve linearly as $z_{k+1}=Az_k+Bu_k$ with a learned $A$ and $B$. The controller solves a QP via \texttt{OSQP} at each step using the learned linear lifted dynamics, with the first-order safety bound $\dot\phi \le b_\phi(x)$ applied linkwise. Figures~\ref{fig:go2_plot_2d}--\ref{fig:go2_linkwise_phi} summarize typical rollouts, showing that the robot follows the reference motion while producing obstacle-avoidance behavior when the safety constraint becomes active.

\subsection*{B. Remarks and Limitations}
Figures~\ref{fig:go2_plot_2d}--\ref{fig:go2_linkwise_phi} demonstrate that the proposed unified Koopman Safe MPC can be applied to a floating-base system when safety is expressed through whole-body kinematics.
Fig.~\ref{fig:title-fig} visualizes the qualitative behavior: the robot tracks the provided motion while producing a clear avoidance maneuver near the obstacle.
Fig.~\ref{fig:go2_plot_2d} aggregates five rollouts and shows that, although the reference passes through the unsafe region, the controller consistently steers the base trajectory around the obstacle.
Fig.~\ref{fig:go2_linkwise_phi} further confirms safety recovery link-by-link: while some links incur brief positive $\phi$ (unsafe-set violations) due to the reference-induced approach, the controller reduces $\phi$ back toward the safe set shortly thereafter.

\smallskip
Whole-body drift is essential.
Unlike fixed-base manipulation, floating-base link motion is strongly influenced by base drift.
Accordingly, the safety derivative must account for the whole-body velocity decomposition $v_{\text{link}}^w=J_b v_{\text{base}}+J_q\dot q$; neglecting the $J_b v_{\text{base}}$ term can under-estimate $\dot\phi$ and delay constraint activation.

\smallskip
Our current implementation evaluates the unit vector $\hat u_d$ and Jacobian blocks $(J_b,J_q)$ at the current measured state and reuses this linearization along the MPC horizon, which yields a single QP per control step.
This approximation is computationally efficient and sufficient to demonstrate feasibility of unified safe control on Go2; however, enforcing time-varying, horizon-accurate constraints would require updating $\hat u_d$ and Jacobians at predicted states.
In addition, since the predictive state omits globally-referenced base translation $(x,y)$, the controller does not explicitly track absolute global position, which contributes to the trajectory variability observed in Fig.~\ref{fig:go2_plot_2d}.
Developing robust horizon-wise geometry updates and global pose propagation (or equivalent local-frame formulations) is left as future work, especially for deployment in real-world environments where global localization may be intermittent.